\documentclass{article} %
\usepackage[final]{colm2025_conference}

\usepackage{microtype}
\usepackage{hyperref}
\usepackage{url}
\usepackage{booktabs}

\usepackage{lineno}

\definecolor{darkblue}{rgb}{0, 0, 0.5}
\hypersetup{colorlinks=true, citecolor=darkblue, linkcolor=darkblue, urlcolor=darkblue}

\usepackage{latexsym}

\usepackage[T1]{fontenc}

\usepackage[utf8]{inputenc}
\usepackage{todonotes}
\usepackage{microtype}

\usepackage{inconsolata}
\usepackage{graphicx}
\usepackage{bm}
\usepackage{svg}
\usepackage{amsmath}
\usepackage{amsfonts}
\usepackage{amssymb}
\usepackage{graphicx}
\usepackage{multirow}
\usepackage{tabularx}
\usepackage{fancyhdr,graphicx,amsmath,amssymb}
\usepackage{wasysym}

\usepackage{algpseudocode}
\usepackage[ruled,linesnumbered,resetcount]{algorithm2e}
\usepackage{multirow}
\usepackage{color}
\usepackage{soul}
\usepackage{graphicx}
\usepackage{tablefootnote}

\usepackage{arydshln}
\usepackage{wrapfig}
\usepackage{listings}
\usepackage{xcolor}

\title{Improving Table Understanding with LLMs and Entity-Oriented Search}

\author{
 \textbf{Thi-Nhung Nguyen\textsuperscript{1}},
 \textbf{Hoang Ngo\textsuperscript{2}},
 \textbf{Dinh Phung\textsuperscript{1}},
 \textbf{Thuy-Trang Vu\textsuperscript{1}},
 \textbf{Dat Quoc Nguyen\textsuperscript{2}}
\\
 \textsuperscript{1}Monash University,
 \textsuperscript{2}Qualcomm AI Research\thanks{Qualcomm Vietnam Company Limited. Qualcomm AI Research is an initiative of Qualcomm Technologies, Inc.}
 \\
 \small{
   \texttt{\{nhung.thinguyen,dinh.phung,trang.vu1\}@monash.edu}, \texttt{\{hoangngo,datnq\}@qti.qualcomm.com}
 }
}

\begin{document}

\ifcolmsubmission
\linenumbers
\fi

\maketitle

\begin{abstract}
Our work addresses the challenges of understanding tables. Existing methods often struggle with the unpredictable nature of table content, leading to a reliance on preprocessing and keyword matching. They also face limitations due to the lack of contextual information, which complicates the reasoning processes of large language models (LLMs). To overcome these challenges, we introduce an entity-oriented search method to improve table understanding with LLMs. This approach effectively leverages the semantic similarities between questions and table data, as well as the implicit relationships between table cells, minimizing the need for data preprocessing and keyword matching. Additionally, it focuses on table entities, ensuring that table cells are semantically tightly bound, thereby enhancing contextual clarity. Furthermore, we pioneer the use of a graph query language for table understanding, establishing a new research direction. Experiments show that our approach achieves new state-of-the-art performances on standard benchmarks WikiTableQuestions and TabFact.
\end{abstract}

\section{Introduction}

\begin{wrapfigure}{r}{0.5\linewidth}
    \centering
\includegraphics[width=0.45\textwidth]{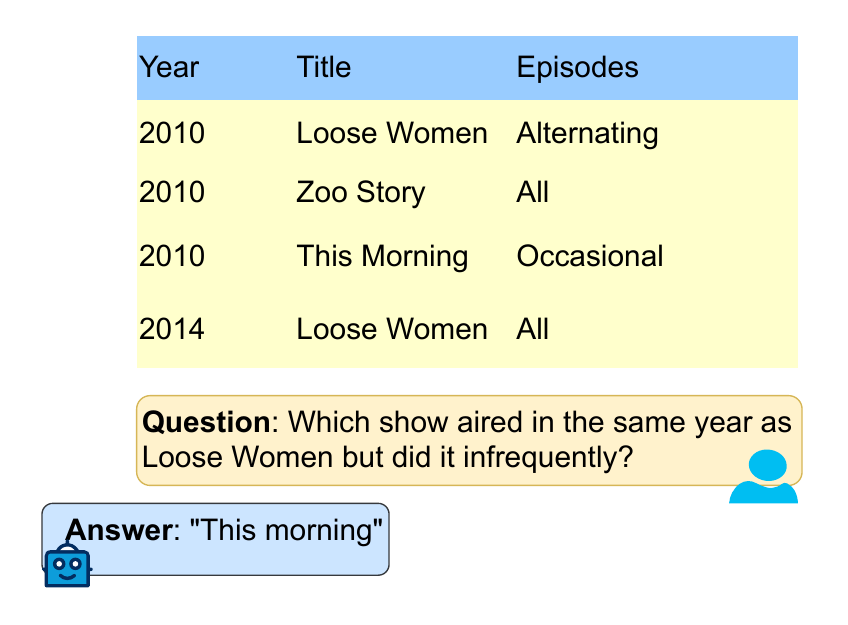}
    \caption{A table question answering example on a "show" table.}
    \label{fig:problem}
\end{wrapfigure}

Tables are widely used to systematically organize and present data. Understanding tables is key to addressing various downstream tasks, such as table-based question answering \citep{wang2023surveytableandtexthybridqaconcepts,lin-etal-2023-inner}. As illustrated in Figure~\ref{fig:problem}, the goal is to extract the relevant information from the table to provide accurate answers to users' questions. 
Recent research has explored using Large Language Models (LLMs) to solve tabular data problems by leveraging their strong performance with prompting \citep{yang-etal-2023-shot,nguyen-etal-2023-self,xie-etal-2023-empirical}. 
One common approach is to convert a natural language question into a structured query (e.g., SQL) and then execute the query on tables to retrieve the final answer \citep{lin-etal-2023-inner, gemmell-dalton-2023-toolwriter, wang2024chainoftable, nahid-rafiei-2024-tabsqlify, liu-etal-2024-rethinking, kong2024opentab}.

Although tabular data allows users to organize and display information logically in real-life scenarios, it presents unique challenges for LLM-based methods. 
One of the major challenges is the unpredictable nature of the content and formatting within table cells. For instance, in a column labeled "\textit{address}", one cell may contain a full address with the street, city, and zip code (e.g., "\textit{123 Main St, Springfield, 12345}"), while another cell may only list the city name (e.g., "\textit{Springfield}") or even be blank. This inconsistency hinders the performance of query search-based methods. To mitigate this challenge, some works have implemented preprocessing techniques, such as normalizing numbers and dates, inputting missing values, removing outliers, and transforming tables into more suitable formats  \citep{gemmell-dalton-2023-toolwriter, zhao-etal-2023-large, liu-etal-2024-rethinking,nahid-rafiei-2024-tabsqlify}. However, these preprocessing techniques demand significant effort in data analysis, are difficult to adapt to unseen tables, and can lead to unintended consequences if not applied properly. Improper preprocessing may result in information loss, creation of sparse matrices, and disruption of the table's original structure.

Another challenge with tabular data is the requirement for complex reasoning due to limited contextual information. This issue arises because table cells often contain only keywords or short phrases rather than full sentences, and the relationships between cells are frequently implicit rather than explicitly stated. For instance, consider a table that includes a "\textit{employee}" column containing names like "\textit{Alice Johnson}" and "\textit{Bob Smith}", along with a "\textit{status}" column indicating their employment status (e.g., "\textit{active}," "\textit{on leave}", or "\textit{terminated}"). Without additional context, interpreting the precise meaning of these statuses can be difficult. To address this challenge, some recent studies have focused on chain-of-thought (COT) reasoning, however, this approach requires multi-step reasoning across the entire table, which can be computationally intensive and resource-demanding \citep{wei2022chain, yao2023tree, chen2023program,  wang2024chainoftable}. Alternative approaches include rephrasing questions, breaking them into sub-queries, or retrieving related columns or rows  \citep{kong2024opentab, patnaik2024cabinet, 10.1145/3539618.3591708}. However, these methods often rely heavily on keyword matching. As a result, the retrieved table cells may be irrelevant to one another, increasing the effort required for LLMs to extract the final answer.

In this paper, we propose a novel approach based on entity-oriented search to enhance table understanding with LLMs. Our method effectively leverages the semantic similarity between the question and the data stored in the table, along with the implicit relationships between cells. Our approach alleviates the strict data format requirements and the reliance on keyword matching found in related works \citep{gemmell-dalton-2023-toolwriter, kong2024opentab, chen2024tablerag}.

Firstly, we focus on identifying entities stored within the table by prompting the LLM. We assume the table contains entities and relations, which may be organized differently depending on the table's structure. An entity can represent a real-world object, person, place, or concept, each defined by its attributes--data cells that provide detailed information. For example, in Figure~\ref{fig:problem}, each show is an entity with attributes like "\textit{Year}", "\textit{Title}", and "\textit{Episodes}". A relationship example is the relation between each entity and its "\textit{Year}", indicating \textit{when a show aired}. By structuring data this way, we aim to establish stronger relationships and constraints among relevant table cells, thereby clarifying the context of each entity. This approach contrasts with the original table data, where excessive information often introduced noise and confusion, making it difficult to determine context.

Secondly, we propose an entity-oriented search approach to extract relevant entities and attributes. Specifically, we integrate methods from full-text search, semantic search, and graph search. Full-text search allows for keyword-based searching, while our semantic search method focuses on the semantic similarities between the entities and attributes in the questions and those stored in the table. Additionally, we use graph queries to represent questions and to query the graphs formed by entities and relations, effectively leveraging these relations to achieve more accurate search results. We expect that this approach will minimize the need for data preprocessing and keyword matching, adapting effectively to various ways of phrasing questions since the underlying meanings and relationships between entities and attributes remain consistent, even when different terms are used. Furthermore, we are pioneering the use of graph query language (Cypher) for table understanding by providing the LLM with both the graph schema and the question as input, enabling it to transform the question into a Cypher query statement.\footnote{\url{https://neo4j.com/docs/Cypher-manual} \citep{10.1145/3183713.3190657}.}

Our contributions are summarized as follows: \textbf{(I)} We propose an entity-oriented search that effectively leverages the semantic similarities between the entities and attributes in the questions and those in the table, along with the implicit relationships between table cells. This minimizes the need for data preprocessing and keyword matching.
\ \ \ \      \textbf{(II)} Our search approach focuses on table entities, ensuring that table cells are semantically tightly bound. This enhances contextual clarity and strengthens relationships between relevant cells.
\ \ \ \      \textbf{(III)}  We are the first to explore a graph query language (Cypher) to enhance table understanding, introducing a new research direction.
\ \ \ \      \textbf{(IV)} Comprehensive experiments show that our approach achieves state-of-the-art performances. %

\section{Related Works}
Table understanding encompasses a wide range of tasks such as question answering (QA). Many early works focus on fine-tuning BERT \citep{kenton2019bert} to serve as a table encoder for these tasks, such as TAPAS \citep{herzig2020tapas}, Table-BERT \citep{Chen2020TabFact}, TABERT \citep{liu2022ptab}, TURL \citep{deng2022turl}, TUTA \citep{wang2021tuta}, and TABBIE \citep{iida2021tabbie}. Recently, the superior performance of large language models (LLMs) with prompting has shifted research focus towards exploring their potential in processing tabular data. A straightforward approach is to concatenate task descriptions with the serialized table as a string and input them into a LLM to generate a text-based response \citep{marvin2023prompt, cheng2023binding, 10.1145/3616855.3635752}. Additionally, some works have further enhanced performance by utilizing few-shot and curated examples \citep{cheng2023binding, narayan2022can, chen2023large}.

To effectively address table-based tasks with large language models (LLMs), recent research increasingly employs external tools instead of relying solely on general text processing. Some works propose generating Python programs and then executing them to extract relevant data \citep{chen2023program, gao2023pal}. Similarly, some works propose using text-to-SQL conversion to extract answers \citep{rajkumar2022evaluating, cheng2023binding, ni2023lever}. However, this approach struggles with complex cases involving intricate tables due to the limitations of the single-pass generation process. In this setup, LLMs cannot modify the table in response to specific questions, necessitating reasoning over a static table. In contrast, a chain-of-thought (CoT)-based approach reasons step by step before providing an answer \citep{chen2023program, zhao2023docmath, yang2024effective}. To enhance CoT, several methods have been proposed, such as breaking down the question into sub-problems \citep{zhou2023leasttomost, khot2023decomposed}, employing a table-filling procedure \citep{ziqi2023tab}, and incorporating preprocessing operations and SQL execution \citep{wang2024chainoftable, nguyen2025planning}.

Additionally, self-consistency (SC), proposed by \citet{wang2023selfconsistency}, is another widely adopted technique in recent state-of-the-art research. SC involves sampling a diverse set of reasoning paths from LLMs and selecting the most consistent answer by marginalizing over these paths \citep{10.1145/3539618.3591708, cheng2023binding, liu-etal-2024-rethinking}. However, a common limitation of both CoT and SC methods is their requirement for a substantial number of reasoning samples from LLMs. For example, \citet{cheng2023binding} generate 50 samples for each question, while \citet{10.1145/3539618.3591708, liu-etal-2024-rethinking} require over 100 samples. This results in slower performance and higher computational costs, making them almost impractical for real-world implementation.

\section{Our Approach}
In this section, we introduce a novel framework, \textbf{TUNES}, to improve \underline{\textbf{t}}able \underline{\textbf{un}}derstanding with \textbf{e}ntity-oriented \textbf{s}earch. Given a table and a related question, the task of TUNES is to generate an answer based on relevant information from the table.

\begin{figure*}[!t]
    \centering
    \includegraphics[width=\textwidth]{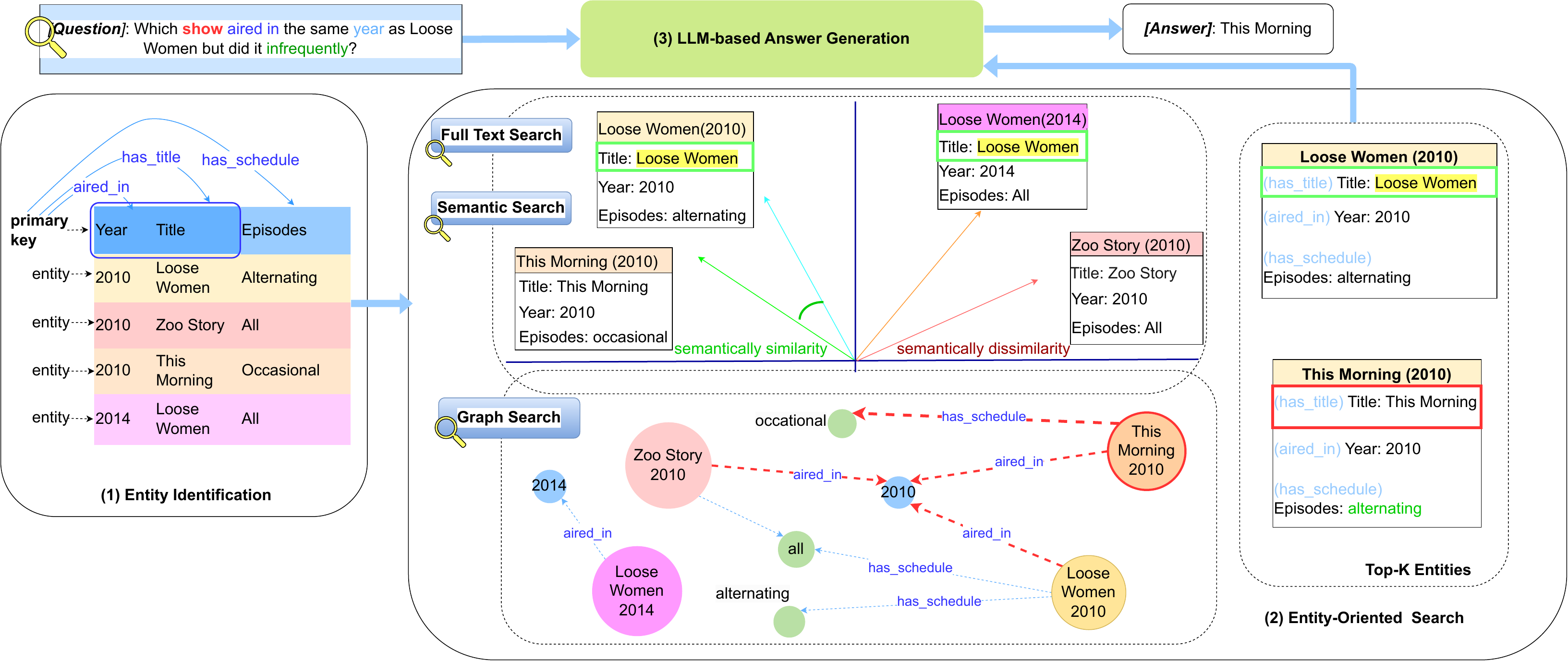}
    \caption{Overview of our proposed framework \textbf{TUNES}. The embedding space and the graph are simplified for illustration purposes.}
    \label{fig:method}
\end{figure*}

\begin{figure*}[t!]
    \centering
\includegraphics[width=0.95\textwidth]{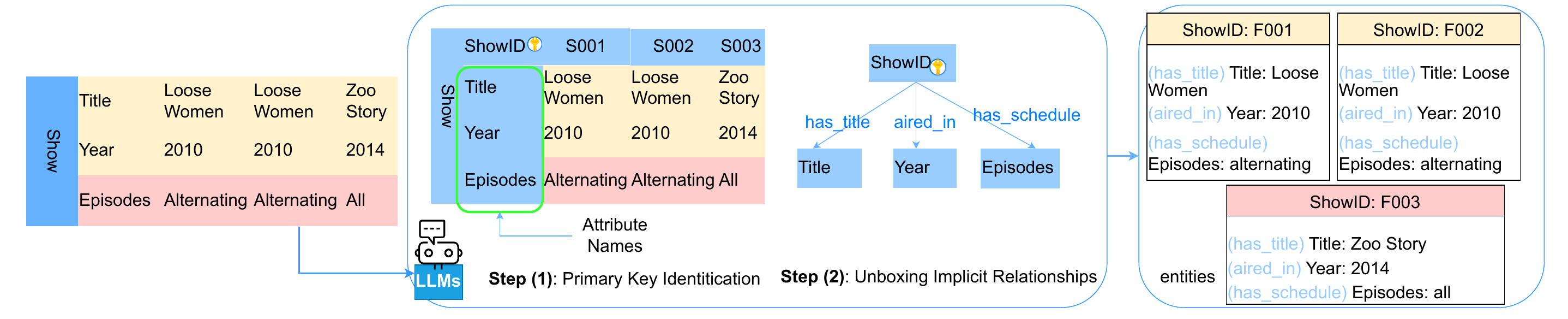}
    \caption{Entity Identification Example.}
    \label{fig:primary_key}
\end{figure*}

Figure~\ref{fig:method} illustrates the architecture of our TUNES, which consists of three main components: (1) Entity Identification, (2) Entity-Oriented Search, and (3) LLM-based Answer Generation. \underline{Entity Identification:} To begin, we focus on extracting entities from the table. We utilize a LLM to analyze the table’s structure, such as primary keys, column names, and row names, to identify attributes and relationships, thereby identifying entities. \underline{Entity-Oriented Search}: We then construct a graph from these entities, attributes, and their relationships. During this process, we remove attributes without values and merge those sharing values to streamline the graph. Simultaneously, we embed the entities, attributes, and user questions into an embedding space, enabling an entity-oriented search that integrates full-text, vector, and graph search. \underline{LLM-based Answer Generation:} Finally, we extract the top K most relevant entities to the question and input them into the LLM to generate accurate answers.

\subsection{Entity Identification}\label{ssection:entityidentification}
This subsection describes our approach to detecting attributes and relationships, which enables the identification of entities within a data table.

The first step is to find the primary key, which can be a single attribute or a combination of attributes that uniquely identifies each entity, typically found in the column or row names. To locate the primary key, we simply use the first $h$ rows and $h$ columns as input to prompt the LLM.  
This process enables us to identify the corresponding attribute column or row and subsequently recognize the entities. For example, in Figure 1, combining "\textit{title}" and "\textit{year}" could serve as the primary key to distinguish the entities within the table. In this case, "\textit{year}," "\textit{title}," and "\textit{episodes}" are attributes, and each entity is identified by its attributes across these columns. Note that, if the primary key is absent from the table, the LLM is responsible for generating it and specifying its position (See our prompt in Table~\ref{tab:prompt-primary-key} in Appendix \ref{sec:appendix-prompt}).

To unbox the implicit relationships in the table, we aim to explore how each entity is related to its attributes. We achieve this by providing the primary key and attribute names to the LLM, prompting it to generate the relationships (See our prompt in Table~\ref{tab:prompt-relations} in Appendix \ref{sec:appendix-prompt}). Figure~\ref{fig:primary_key} illustrates an example of the entity identification process. 

\subsection{Entity-Oriented Search}\label{ssection:hybrid}

\paragraph{Graph Search:} For each table, we construct a graph $\mathcal{G} = {\mathcal{N}, \mathcal{E}}$, where $\mathcal{N}$ is the set of nodes and $\mathcal{E}$ is the set of edges.   $\mathcal{N}$ represents the entities and attributes stored in the table, formed by the union of entity nodes and attribute nodes. To preserve the position of nodes in the table, we store their table addresses as properties of each node. $\mathcal{E}$ represents the relationships within the table, established by the relationships between the entity and its attributes. 

\begin{figure*}[t!]
    \centering
\includegraphics[width=0.9\textwidth]{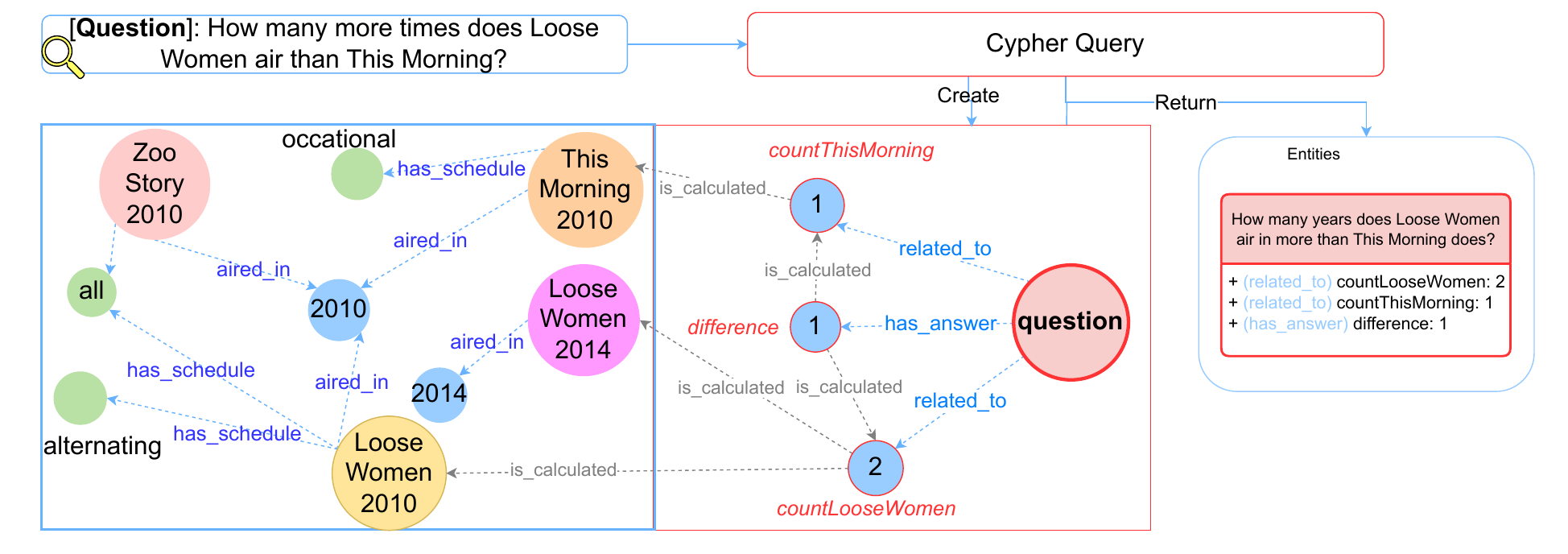}
    \caption{Cypher Query Execution Process. To answer the question "How many more times does Loose Women air than This Morning?", the Cypher query execution process first calculates the number of times "Loose Women" airs, creating a new attribute called "countLooseWomen". It then performs a similar calculation for "This Morning", generating an attribute called "countThisMorning". The process calculates the difference between "countLooseWomen" and "countThisMorning", creating an attribute named "difference". Note that the relationship "is\_calculated" is used solely to illustrate the calculation process. Next, the process creates a new entity representing the question and determines how this entity relates to the newly created attributes, using relationships such as "related\_to" and "has\_answer". Finally, the process returns a complete entity that effectively answers the question by providing a comprehensive analysis of the attributes' impact. See the corresponding Cypher query for the input question in Appendix~\ref{sec:cypher-example}. \textit{We pioneer the use of the graph query language Cypher to improve table understanding.}}
    \label{fig:Cypher}
\end{figure*}

\underline{Node disambiguation:}\ \ We exclude attributes that lack values, which results in sparse data in the original table. Also, some attributes may refer to the same value or have synonymous meanings. For example, in Table~\ref{fig:problem}, cells (1,0), (2,0), and (3,0) all correspond to the year "\textit{2010}", and these should be merged into a single node. We accomplish this by filtering based on the semantic similarity score between two node names in an embedding space, as defined in the next paragraph. To query the graph, we provide the LLM with both the graph schema and the question as input, prompting it to convert the question text into a Cypher query statement (See our prompt in Table~\ref{tab:prompt-text-to-Cypher} in Appendix \ref{sec:appendix-prompt}). %
We then execute this Cypher query on the graph to search relevant entities and attributes. %
In addition to searching for relevant entities and verifying the constraints that must be satisfied in questions, the Cypher query execution process enables complex calculations on attributes and entities. It allows for the automatic generation of new entities along with their associated attributes, as illustrated in Figure~\ref{fig:Cypher}, making graph search a crucial component of TUNES.

\paragraph{Semantic Search:} While constructing the graph, we simultaneously map both the entity nodes and the user's question into an embedding space using a text embedding model. This is to calculate the semantic search scores by determining the similarity between the question and the entities. We measure this similarity using the cosine similarity between their representation vectors. Specifically, we obtain the question's representation by directly inputting the question text into the text embedding model. For each entity node, we generate a representation by feeding the embedding model a concatenation of node names from a sub-graph of depth $d$, with the entity node as the root.

\paragraph{Full-text Search:} Full-text search determines relevance by considering the number of keyword matches and their frequency. We use the BM25 algorithm to rank table entities based on how well they align with the given question \citep{10.1561/1500000019}. First, we create a search document for each entity by combining its primary key and attributes. Then, we calculate a search score for each entity using BM25, comparing the question to its search document, with higher scores indicating a stronger match to the question.

\paragraph{Entity-Oriented Search Approach:} Entity-Oriented search combines full-text search, semantic search, and graph search to provide more relevant search results. Specifically, we extract the top K relevant entities by calculating a weighted sum of their full-text and semantic search scores, while also retrieving entities and attributes output from the Cypher query execution process. Full-text search quickly identifies exact terms or closely related ones, even in large datasets. Semantic search focuses on understanding the underlying context or meaning of the question.  Graph search addresses constraints and implicit relationships within the question. This approach ensures that even if a question does not precisely match a relevant entity, the relevant meanings and satisfied relational constraints within the graph will still allow it to be retrieved.

\subsection{LLM-based Answer Generation}
The top K relevant entities are incorporated into a prompt and input into the LLM to generate a response (See our prompt in Table~\ref{tab:prompt-answer} in Appendix \ref{sec:appendix-prompt}). Each entity is treated as a paragraph, with the primary key serving as a heading that introduces the main topic, while the attributes and relations provide further details and elaboration. Combining these entities is akin to constructing a document, where each paragraph, representing an entity, can stand alone but may also connect to other paragraphs if two entities share the same attribute. This approach aligns well with LLMs, which are extensively trained on text documents, reducing the need for complex reasoning and resulting in a more accurate answer.

\section{Experiment Setup}

\paragraph{Datasets \& Metric:}

Following previous works \citep{wang2024chainoftable}, we conduct experiments on the benchmark datasets WikiTableQuestions---a question answering dataset over semi-structured tables \citep{pasupat2015compositional} and TabFact---a dataset for table-based fact verification \citep{Chen2020TabFact}. See Appendix \ref{appendix:datasets} for the statistics of their test sets.  

We employ the binary classification accuracy for TabFact and the official denotation accuracy for WikiTableQuestions \citep{pasupat2015compositional}.

\paragraph{Baselines:}

We compare TUNES with strong baselines, including: \textbf{(I)} Methods based on self-consistency (SC) or chain-of-thought (CoT), which require a substantial number of reasoning inferences with LLMs: \textbf{Dater} \citep{khot2023decomposed}, \textbf{BINDER} \citep{cheng2023binding},  \textbf{CHAIN-OF-TABLE} \citep{wang2024chainoftable} and \textbf{DP\&PYAGENT} \citep{liu-etal-2024-rethinking}.
\textbf{(II)} Methods without SC and CoT, which only require a few number of LLM inferences: \textbf{TEXT2SQL} \citep{rajkumar2022evaluating}, \textbf{ChatGPT} \citep{jiang2023structgpt}, \textbf{StructGPT} \citep{jiang2023structgpt}, \textbf{TableRAG} \citep{chen2024tablerag} and \textbf{TabSQLify}  \citep{nahid-rafiei-2024-tabsqlify}.

\paragraph{Implementation Details:} 
Following previous works, we use \textit{GPT-3.5-turbo} as the LLM. Additionally, we report our scores with other LLMs, including \textit{GPT-4o-mini} and the open-weight LLMs \textit{Llama-3.1-8B-Instruct} and \textit{Llama-3.1-70B-Instruct} \citep{dubey2024llama}. We use the OpenAI API to run inferences with \textit{GPT-3.5-turbo} and \textit{GPT-4o-mini}, while vLLM \citep{kwon2023efficient} is used for inference with \textit{Llama-3.1-8B-Instruct} and \textit{Llama-3.1-70B-Instruct}. For all these models, we set the temperature values to 0.4 for Text-to-Cypher generation and 0.0 for answer generation. %

To extract primary keys (see Section \ref{ssection:entityidentification}), we set $h$ to 5. We use Neo4J to interact with graphs (see Section \ref{ssection:hybrid}).\footnote{\url{https://neo4j.com/docs/Cypher-manual}} We merge attribute nodes through exact matching and filter them by a cosine similarity greater than 0.95. For text embeddings, we use \textit{bge-m3} \citep{chen2024bge} as our embedding model, setting $d = 2$ when embedding nodes (see Section \ref{ssection:hybrid} - Semantic Search). 
For the entity-oriented search approach (see Section \ref{ssection:hybrid}), the weight of each search component is set to 1, and for the top K most relevant entities, we set K to 50.

\paragraph{TUNES [CoT]:} We also develop a variant of \textbf{TUNES with CoT}. Specifically, we require the model to answer the question step by step by iteratively executing the search and answer processes. Simultaneously, the LLM-based Answer Generation is required to reason step by step before delivering the final answer. The maximum number of iterations is set to 3.

\begin{table}[!t]
\centering
\begin{tabular}{l|l|c|c}
\hline
\multicolumn{2}{c|}{\textbf{Approach}} & \textbf{WikiTQ} & \textbf{TabFact} \\ \hline 
\multirow{10}{*}{\rotatebox[origin=c]{90}{{GPT-3.5-turbo}}} & Dater \citep{khot2023decomposed} \textbf{[SC]} (*) & 52.8  & 78.0 \\
& BINDER \citep{cheng2023binding} \textbf{[SC]} (*) & 56.7 & 79.2 \\
& CHAIN-OF-TABLE \citep{wang2024chainoftable} \textbf{[CoT]} (*)  & 59.9 & {80.2} \\
& DP\&PYAGENT \citep{liu-etal-2024-rethinking} \textbf{[SC]} (*)  & {65.5} & 80.0 \\ 
& Our TUNES\textsubscript{GPT-3.5-turbo} \textbf{[CoT]}& \textbf{68.5} & \textbf{81.5} \\ \cline{2-4} 
& StructGPT \citep{jiang2023structgpt} & 48.4 & \_\\
& ChatGPT \citep{jiang2023structgpt} & 51.8 & \_ \\
& TableRAG \citep{chen2024tablerag}  & 57.0 & \_ \\
& TEXT2SQL \citep{rajkumar2022evaluating} & 52.9 & 64.7 \\
& TabSQLify \citep{nahid-rafiei-2024-tabsqlify} & {64.7} & {79.5} \\ %
& Our TUNES\textsubscript{GPT-3.5-turbo} & \textbf{64.9} & \textbf{79.8} \\ \hline \hline
\multirow{12}{*}{\rotatebox[origin=c]{90}{{Others.}}}  
& CHAIN-OF-TABLE\textsubscript{GPT-4o-mini} \textbf{[CoT]} (*) & 70.4 & 85.8 \\
& DP\&PYAGENT\textsubscript{GPT-4o-mini} \textbf{[SC]} (*) & 74.7 & 89.9 \\
& Our TUNES\textsubscript{GPT-4o-mini} & {72.3} & {84.7} \\
& Our TUNES\textsubscript{GPT-4o-mini} \textbf{[CoT]} & \textbf{75.4} & \textbf{90.4}  \\
\cline{2-4}
& CHAIN-OF-TABLE\textsubscript{Llama-3.1-70B-Instruct} \textbf{[CoT]} & 70.1 & 85.6 \\
&  DP\&PYAGENT\textsubscript{Llama-3.1-70B-Instruct} \textbf{[SC]} & 67.9 & 85.1 \\
& Our TUNES\textsubscript{Llama-3.1-70B-Instruct} & 75.4 & 85.6 \\ 
& Our TUNES\textsubscript{Llama-3.1-70B-Instruct} \textbf{[CoT]} & \textbf{75.7} & \textbf{87.5}  \\
\cline{2-4}
& CHAIN-OF-TABLE\textsubscript{Llama-3.1-8B-Instruct} \textbf{[CoT]} & 56.0 & 49.6 \\
& DP\&PYAGENT\textsubscript{Llama-3.1-8B-Instruct} \textbf{[SC]} & 57.3 & 63.8 \\
& Our TUNES\textsubscript{Llama-3.1-8B-Instruct} & {54.1} & {68.1} \\
& Our TUNES\textsubscript{Llama-3.1-8B-Instruct} \textbf{[CoT]} & \textbf{57.8} & \textbf{71.9}  \\
\hline
\end{tabular}
    \caption{Performance results on the WikiTableQuestions (WikiTQ) and TabFact test sets. [SC] and [CoT] denote approaches based on self-consistency and  chain-of-thought, respectively. In rows 2--15, results for previous methods are taken from their respective works, except for Dater, BINDER, CHAIN-OF-TABLE, and DP\&PYAGENT, which are marked with (*), are taken from \cite{nguyen2025planning}.
    In the last 8 rows, we run the official implementations of CHAIN-OF-TABLE (\url{https://github.com/google-research/chain-of-table}) and DP\&PYAGENT (\url{https://github.com/Leolty/tablellm}) using Llama-3.1-8B-Instruct and Llama-3.1-70B-Instruct.}

\label{tab:main_result}
\end{table}

\section{Results}

\subsection{Main Results}
\label{sec:result}

Using GPT-3.5-turbo as the base LLM, as shown in Table \ref{tab:main_result},  when compared to baselines employing SC and CoT, our TUNES\textsubscript{GPT-3.5-turbo} [CoT] outperforms the previous state-of-the-art (SOTA) model DP\&PYAGENT by 3.0\% points on WikiTableQuestions and 1.5\% on TabFact. Notably, TUNES\textsubscript{GPT-3.5-turbo} [CoT] requires fewer intermediate responses from the LLM (the total is \underline{8}, including 2 for table analysis, 3 for searching and 3 for answering), compared to CHAIN-OF-TABLE (\underline{25}), Binder (\underline{50}), Dater (\underline{100}) \citep{wang2024chainoftable}, and DP\&PYAGENT (\underline{50}-\underline{150}) \citep{liu-etal-2024-rethinking}. Note that the average time for performing both semantic search and full-text search is very small, only at 0.06 seconds per query on a CPU. Thus, in TUNES, almost all of the running time is spent on prompting LLMs. As a result, using entity-oriented search has allowed us to reduce the LLM inference cost per question by a factor of $25  / 8 \simeq 3$ to $150 / 8 \simeq 18$. In addition, our original TUNES\textsubscript{GPT-3.5-turbo} surpasses all baselines not utilizing CoT and SC, outperforming TEXT2SQL by 12\% on WikiTableQuestions and 15.1\% on TabFact, while achieving a 0.2+\% improvement over the previous SOTA model TabSQLify on both datasets. 

In the last 12 rows of Table \ref{tab:main_result}, 
we benchmark TUNES against SOTA methods, including CHAIN-OF-TABLE and DP\&PYAGENT, across different LLMs, including Llama-3.1-8B-Instruct, Llama-3.1-70B-Instruct, GPT-4o-mini to further evaluate the adaptability of our approach. The results show that the original TUNES demonstrates competitive performance, while TUNES [CoT] consistently outperforms both CHAIN-OF-TABLE and DP\&PYAGENT across all examined LLMs and datasets.  

The performance of  TUNES\textsubscript{GPT-4o-mini} remains consistent with the results previously observed on GPT-3.5-turbo. In detail, compared to DP\&PYAGENT\textsubscript{GPT-4o-mini} [SC], TUNES\textsubscript{GPT-4o-mini} is 2.4\% points lower on WikiTQ and 5.2\% points lower on TabFact. However, TUNES\textsubscript{GPT-4o-mini} surpasses CHAIN-OF-TABLE\textsubscript{GPT-4o-mini} [CoT] on WikiTQ by 1.9\% points, despite being 1.1\% points lower on TabFact.  When augmented with CoT reasoning, TUNES\textsubscript{GPT-4o-mini} [CoT] surpasses both baselines, exceeding CHAIN-OF-TABLE\textsubscript{GPT-4o-mini} [CoT] by 5.0\% points on WikiTQ and 4.6\% points on TabFact, while also outperforming DP\&PYAGENT\textsubscript{GPT-4o-mini} [SC] by 0.7\% points on WikiTQ and 0.5\% points on TabFact.

For open-source LLMs, TUNES demonstrates better adaptability than the baselines. TUNES\textsubscript{Llama-3.1-8B-Instruct} notably outperforms CHAIN-OF-TABLE\textsubscript{Llama-3.1-8B-Instruct} [CoT] by 18.5\% points and DP\&PYAGENT\textsubscript{Llama-3.1-8B-Instruct} [SC] by 4.3\% points on TabFact. In addition, TUNES\textsubscript{Llama-3.1-70B-Instruct} even surpass both baselines on both datasets, demonstrating a 5.3\% points improvement on WikiTQ over CHAIN-OF-TABLE\textsubscript{Llama-3.1-70B-Instruct} [CoT] and 7.5\% points on WikiTQ and 0.5\% on Tabfact over DP\&PYAGENT\textsubscript{Llama-3.1-70B-Instruct} [SC]. Meanwhile, TUNES\textsubscript{Llama-3.1-8B-Instruct} [CoT] and TUNES\textsubscript{Llama-3.1-70B-Instruct} [CoT] consistently achieve SOTA performance on both datasets. All obtained results demonstrate the effectiveness of our proposed approach. 

Overall, TUNES without CoT requires 6× to 36× fewer LLM calls than the baselines, resulting in significant computational savings. Despite this efficiency, TUNES still achieves competitive performance across four different LLMs, demonstrating its generalizability and effectiveness. Meanwhile,  when combined with CoT, TUNES requires 3×–18× fewer LLM calls, while significantly outperforming state-of-the-art baselines such as CHAIN-OF-TABLE and DY\&PYAGENT, with statistical significance (p < 0.01)\footnote{Based on one-sided McNemar tests from in-house experiments.} on both the WikiTQ and TabFact datasets.

\begin{wraptable}{r}{0.5\linewidth}
\centering
\begin{tabular}{l|c}
\hline
\textbf{Approach} & \textbf{WikiTQ} \\ \hline 
TUNES\textsubscript{GPT-4o-mini} & 72.0 \\ \hdashline %
\ \ \ \ without Entity Identification & 61.3 \\
\ \ \ \ without Graph Search & 62.0 \\ %
\ \ \ \ without Semantic Search & 69.7 \\ %
\ \ \ \ without Full-text Search & 70.2 \\
\hline
\end{tabular}
\caption{Ablation results for TUNES\textsubscript{GPT-4o-mini}.}
\label{tab:ablation_study}
\end{wraptable}

\subsection{Ablation Study} \label{ablation_study}

To assess the impact of each proposed component of TUNES, we conduct an ablation study by evaluating different ablated versions of TUNES\textsubscript{GPT-4o-mini}. Given budget constraints, we evaluate these ablated variants on a randomly selected subset of \textbf{1,000} questions from the WikiTableQuestions test set. Table \ref{tab:ablation_study} presents the results for the full-component TUNES\textsubscript{GPT-4o-mini} alongside the ablation study results on this subset.

\textbf{Without Entity Identification:} We exclude the entity identification component from TUNES. Instead of searching for entities, the search strategy shifts from an entity-oriented approach to a row-oriented one. That is, each row becomes a search document for both full-text and semantic searches and serves as a node in the graph without any relationships. The results indicate that this shift from entity-oriented to row-oriented search reduces accuracy, with a drop of 72.0\% - 61.3\% = 10.7\% compared to the full TUNES. This shows that clarifying the context for table entities, along with the relationships between entities and their attributes, enhances performance, underscoring the effectiveness of the proposed approach.

\textbf{Without Graph Search:} Graph search is excluded from the search strategy, leaving the task of searching solely to full-text and semantic searches. As shown in Table~\ref{tab:ablation_study}, the removal of graph search reduces performance, resulting in a 10\% decrease in accuracy. This decrease shows that utilizing Cypher, a graph query language, to query the graph effectively leverages the relationships between entities and attributes, along with the constraints that must be satisfied in the question. As a result, it produces more relevant entities in the search results.%

\textbf{Without Semantic Search:} Semantic search is excluded from the search strategy. As shown in Table~\ref{tab:ablation_study}, removing semantic search decreases TUNES's performance by 2.3\%.

\textbf{Without Full-Text Search:} Full-text search is excluded from the search strategy. Although this removal negatively impacts TUNES's performance, the decrease of 1.8\% is not as large as with the removal of other components. This shows that the integration of graph queries and semantic search in TUNES reduces reliance on keyword matching.

\begin{table*}[!t]
\centering
\resizebox{14cm}{!}{
\begin{tabularx}{\linewidth}{>{\hsize=0.3\hsize}X >{\hsize=0.5\hsize}X >{\centering\arraybackslash}p{6mm} >{\centering\arraybackslash}p{5mm}}
\hline
\textbf{Error Type} & \textbf{Description} & \textbf{\%} & \textbf{Exa.} \\ \hline
\textbf{[Entity]} Table structure identification & The LLM incorrectly identifies the positions of primary keys and attributes. & 4\% & \ref{sec:error-entity}\\ \hline
\textbf{[Search]} Insufficient entity retrieval & Entity-oriented search fails to retrieve an adequate number of entities required to answer the question. & 8\% & \ref{sec:error-search-insufficient} \\ 
\textbf{[Search]} Incorrect self-generated entity & Cypher execution process generates inaccurate information. & 50\% & \ref{sec:error-search-incorrect} \\ \hline
\textbf{[Answer]} Comparative error & The LLM incorrectly compares quantities such as distance or time. & 8\% & \ref{sec:error-answer-comparative} \\ 
\textbf{[Answer]} Numerical error & The LLM performs incorrect calculations. & 24\% & \ref{sec:error-answer-numeric} \\ 
\textbf{[Answer]} Logical error & The LLM incorrectly extracts and utilizes provided information in the reasoning process. & 20\% & \ref{sec:error-answer-logical} \\ 
\textbf{[Answer]} Others & N/A & 2\% & \\ \hline
\end{tabularx}
}
\caption{Error types by components---Entity Identification (denoted as [Entity]), Entity-Oriented Search (denoted as [Search]), and LLM-based Answer Generation (denoted as [Answer])---in TUNES\textsubscript{Llama-3.1-70B-Instruct}. The total percentage does not add up to 100\% as some samples contain more than one error. See error examples (Exa.) in Appendix \ref{sec:example-error}.}
\label{tab:error}
\end{table*}

\subsection{Error Analysis}
\label{error_analysis}

Table \ref{tab:error} reports the types of errors across each component from TUNES\textsubscript{Llama-3.1-70B-Instruct} on WikiTableQuestions. 

\textbf{Entity Identification Errors:}\ \  The LLM excels in entity identification within tables, maintaining an error rate of just 4\%. Most errors occur in tables with complex structures, especially those with duplicated row and column names.

\textbf{Entity-Oriented Search Errors:}\ \ Here, 58\% of the errors are related to the quality of the entity-oriented search. Among these, only 8\% are due to an insufficient number of retrieved entities, which can result from Cypher query syntax errors or questions requiring a large number of entities. The main issue lies in our Cypher execution process, which generates inaccurate new entities and attributes due to mistakes in intermediate calculations, such as incorrectly selecting entities or calculating functions.

\textbf{ LLM-based Answer Generation Errors:}\ \ The biggest challenge for the LLM is performing calculations, such as addition and subtraction, which have an error rate of 24\%. The second major challenge is logical errors, with an error rate of 20\%. These errors occur because the LLM does not fully understand the question or the table's content, leading to incorrect information extraction. Other challenges include errors in comparing complex quantities, such as determining which athlete finishes a race the fastest.

Overall, TUNES still faces challenges related to inaccurate information generated from Cypher queries and the limitations of the LLM in calculations, comparisons, and reasoning.

\section{Conclusion}

We propose a novel approach \textbf{TUNES} to tackle the challenges of table understanding, with three main goals: (1) effectively leveraging the semantic similarities between questions and table data, along with the implicit relationships between table cells, to reduce the need for data preprocessing and keyword matching; (2) ensuring that table cells are semantically tightly connected to enhance contextual clarity; and (3) pioneering the use of a graph query language (Cypher) to improve table understanding. 
Experimental results show that TUNES achieves a state-of-the-art performance. In the future, we plan to extend TUNES to address other complex downstream tasks related to table understanding. Our TUNES implementation is publicly available at: \texttt{https://github.com/nhungnt7/TUNES}.

\section*{Acknowledgement}
This work was supported by Monash eResearch capabilities, including M3. 

This work was completed while Hoang Ngo and Dat Quoc Nguyen were at Movian AI, Vietnam.

\bibliography{colm2025_conference}
\bibliographystyle{colm2025_conference}

\newpage 

\appendix

\section{Prompts}
\label{sec:appendix-prompt}

Tables \ref{tab:prompt-primary-key}, \ref{tab:prompt-relations}, \ref{tab:prompt-text-to-Cypher} and \ref{tab:prompt-answer} show prompts used in our TUNES framework. 

\begin{table*}[]
\begin{tabularx}{\linewidth}{X}
\hline
\textbf{Prompt} \\ \hline
Task: Given a table, your task is to identify the primary key and its position, and then identify the position of the attribute names within the table.\\ 
 
If the table does not contain a primary key, generate one and return its position. The table positions should be referenced as a 2D Python array, with indexing starting at [0, 0]. Negative indices, such as -1 or -2, may be used for the inserted primary key.\\ 
 
Output Template:\\ 
Primary Key: [<a list of primary key>]\\ 
Primary Key Position: \{'column': [<column numbers>]\} or \{'row': [<row numbers>]\}\\ 
Attribute Names Position: column or row\\ 
 
Examples:\\ 
\{example\}\\ 
 
Complete:\\ 
\{table\}\\ 
Output:\\
\hline
\end{tabularx}
\caption{Prompt to identify primary key.}
\label{tab:prompt-primary-key}
\end{table*}

\begin{table*}[]
\begin{tabularx}{\linewidth}{X}
\hline
\textbf{Prompt} \\ \hline
Your task is to find the relationship between primary\_key and attributes, along with a description.\\ 
 
Output Template:\\ 
- <attribute name>: <relationship with primary key> | description\\ 
 
Input and Output example:\\ 
\{examples\}\\ 
 
Complete:\\ 
attributes: \{attributes\}\\ 
primary\_key: \{primary\_key\}\\ 
relationships:\\ 

\hline
\end{tabularx}
\caption{Prompt to generate relations.}
\label{tab:prompt-relations}
\end{table*}
\begin{table*}[]
\begin{tabularx}{\linewidth}{X}
\hline
\textbf{Prompt} \\ \hline
Schema:\\ 
\{schema\}\\ 
Your task is to extract a subgraph containing all necessary nodes to answer to question. Please return Cypher code only.\\ 
Note that: \\ 
1. Value of all properties is string (convert to number if needed). \\ 
2. If the question is to the order of the value in the table, please answer based on the properties column\_address (int) and row\_address (int) of the table.\\ 
3. If the question requires compare strings please use toLower to compare both.\\ 
 
Input and Output example:\\ 
\{examples\}\\ 
 
Complete:\\ 
Question: \{question\}\\ 
Cypher code:\\ 
\hline
\end{tabularx}
\caption{Prompt to generate Cypher query.}
\label{tab:prompt-text-to-Cypher}
\end{table*}
\begin{table*}[]
\begin{tabularx}{\linewidth}{X}
\hline
\textbf{Prompt} \\ \hline
Context:\\ 
\{context\}\\ 
Question:\\ 
\{question\}\\ 
Your task is to answer the question based on given context. Please provide the answer extracted only, do not include any rewrite or new content.\\ 
If the question is related to the location of the data in the table, please answer based on the column address or row address. Note that -1 mean all the column/row.\\ 
You can explain the answer in a short context in the next row and show the confidence score.\\ 
\hline
\end{tabularx}
\caption{Prompt to generate the final answer.}
\label{tab:prompt-answer}
\end{table*}

\section{Cypher code example}
\label{sec:cypher-example}
Cypher code to retrieve data for query "How many more times does Loose Women air than This Morning?" is shown in Figure~\ref{fig:cypher-example}.

\lstset{
    language=SQL, %
    morekeywords={MATCH, WITH, CREATE, RETURN,  Entity, Attribute}, %
    basicstyle=\ttfamily\small,
    keywordstyle=\color{blue}\bfseries,
    commentstyle=\color{green},
    stringstyle=\color{red},
    showstringspaces=false,
    numbers=none,
    numberstyle=\tiny,
    breaklines=true,
    frame=single,
}
\begin{figure*}[!t]
\centering
\begin{lstlisting}
MATCH (:Entity {title: 'Loose Women'})-[:air_in]->(y:Year)
WITH collect(y.value) AS yearsLooseWomen
MATCH (:Entity {title: 'This Morning'})-[:air_in]->(y:Year)
WITH yearsLooseWomen, collect(y.value) AS yearsThisMorning
WITH size(yearsLooseWomen) as countLooseWomen,
     size(yearsThisMorning) as countThisMorning,
     size(yearsLooseWomen) - size(yearsThisMorning) as difference

CREATE (lw: Attribute {countLooseWomen: countLoosewomen})
CREATE (tm: Attribute {countThisMorning: countThisMorning})
CREATE (diff: Attribute {difference: difference})
CREATE (result: Entity {query: "How many times does Loose Women air in more than This Morning?"})
CREATE (result)-[:related_to]->(lw)
CREATE (result)-[:related_to]->(tm)
CREATE (result)-[:has_answer]->(diff)

RETURN result, lw, tm, diff
\end{lstlisting}
\caption{Cypher code example to retrieve data for query "How many more times does Loose Women air than This Morning?"}
\label{fig:cypher-example}
\end{figure*}

\section{Dataset statistics}\label{appendix:datasets}

Table \ref{tab:dataset_statistic} presents the statistics of the WikiTableQuestions and TabFact test sets.  

\begin{table}[ht!]
\centering
\resizebox{7cm}{!}{
\begin{tabular}{l|l|l}
\hline
\textbf{Statistics} & \textbf{WikiTQ}  &  \textbf{TabFact}\\ \hline 
Number of Questions & 4343 & 2024 \\
Number of Tables & 421  & 298 \\
Min/Max Number of Rows & 6/518 & 5/49 \\
Min/Max Number of Columns & 5/20 & 3/21 \\ \hline
\end{tabular}
}
\caption{Statistics of the WikiTableQuestions (WikiTQ) and TabFact test sets.}%
\label{tab:dataset_statistic}
\end{table}

\section{Example of errors}
\label{sec:example-error}
\subsection{Entity identification error}
\subsubsection{Table structure identification error}
\begin{figure*}
    \centering
    \includegraphics[width=\linewidth]{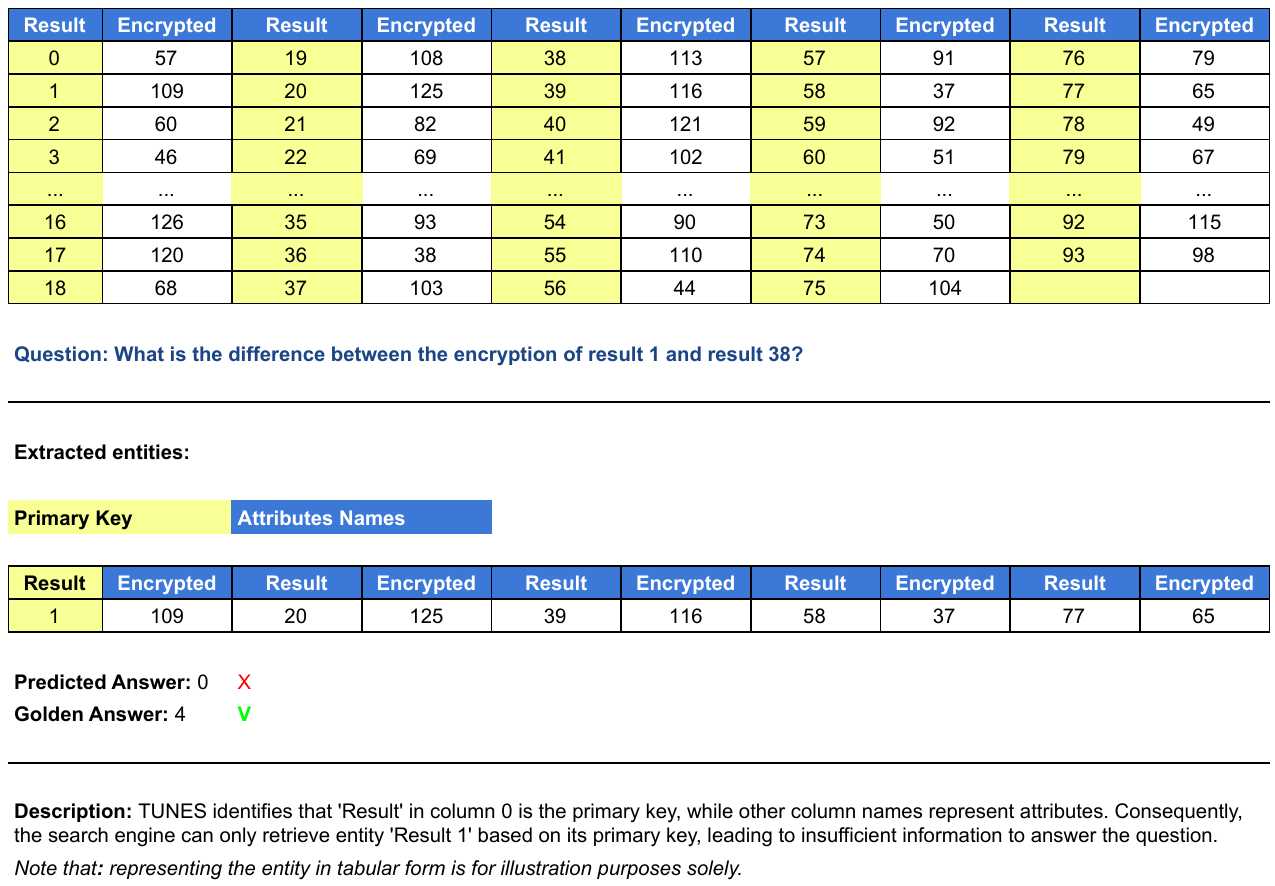}
    \caption{Illustration of Table structure identification error.}
    \label{fig:entity-error}
\end{figure*}

Figure~\ref{fig:entity-error} illustrates a table structure identification error.
\label{sec:error-entity}

\subsection{Search errors}
\subsubsection{Insufficient entity retrieval error}
\begin{figure*}
    \centering
    \includegraphics[width=\linewidth]{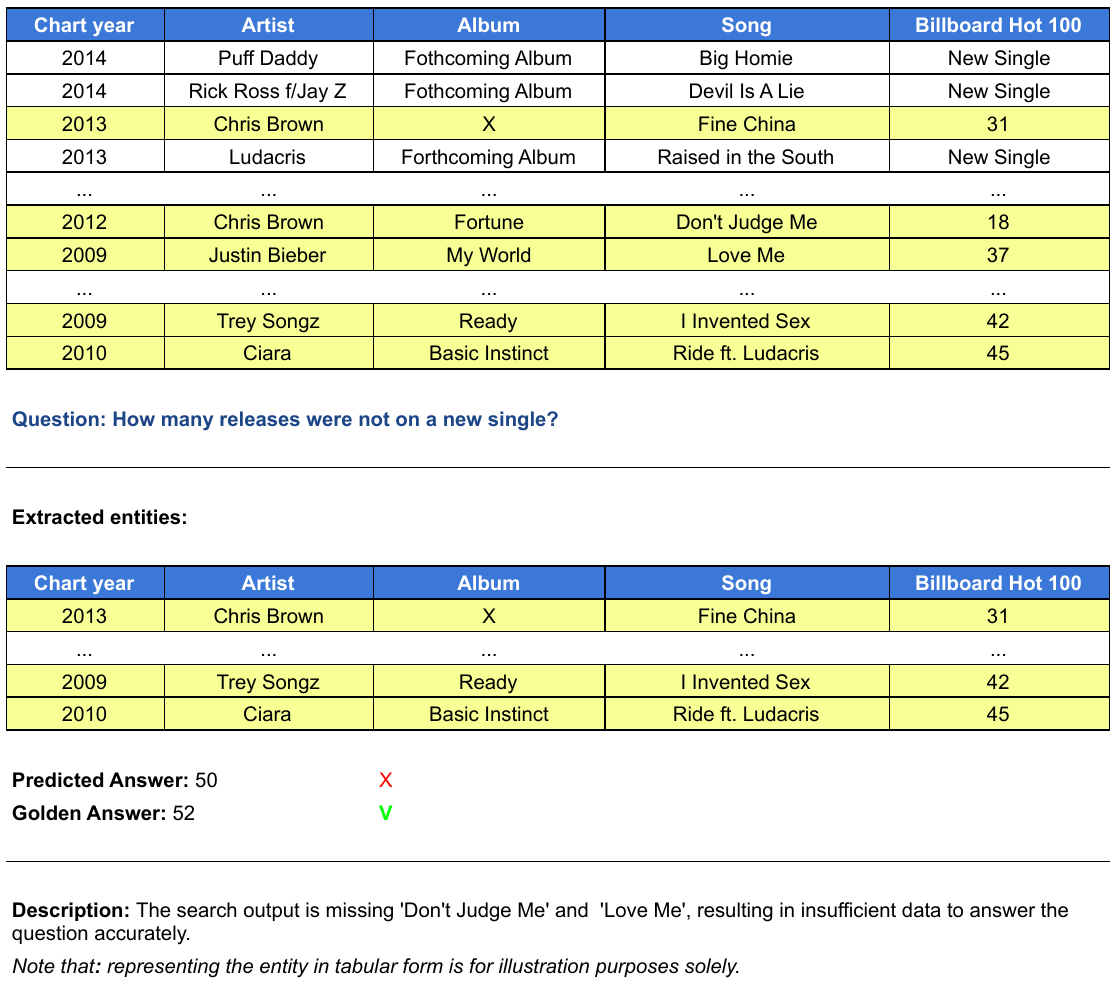}
    \caption{Illustration of Insufficient entity retrieval error.}
    \label{fig:fullttext-search-error}
\end{figure*}

\begin{figure*}
    \centering
    \includegraphics[width=\linewidth]{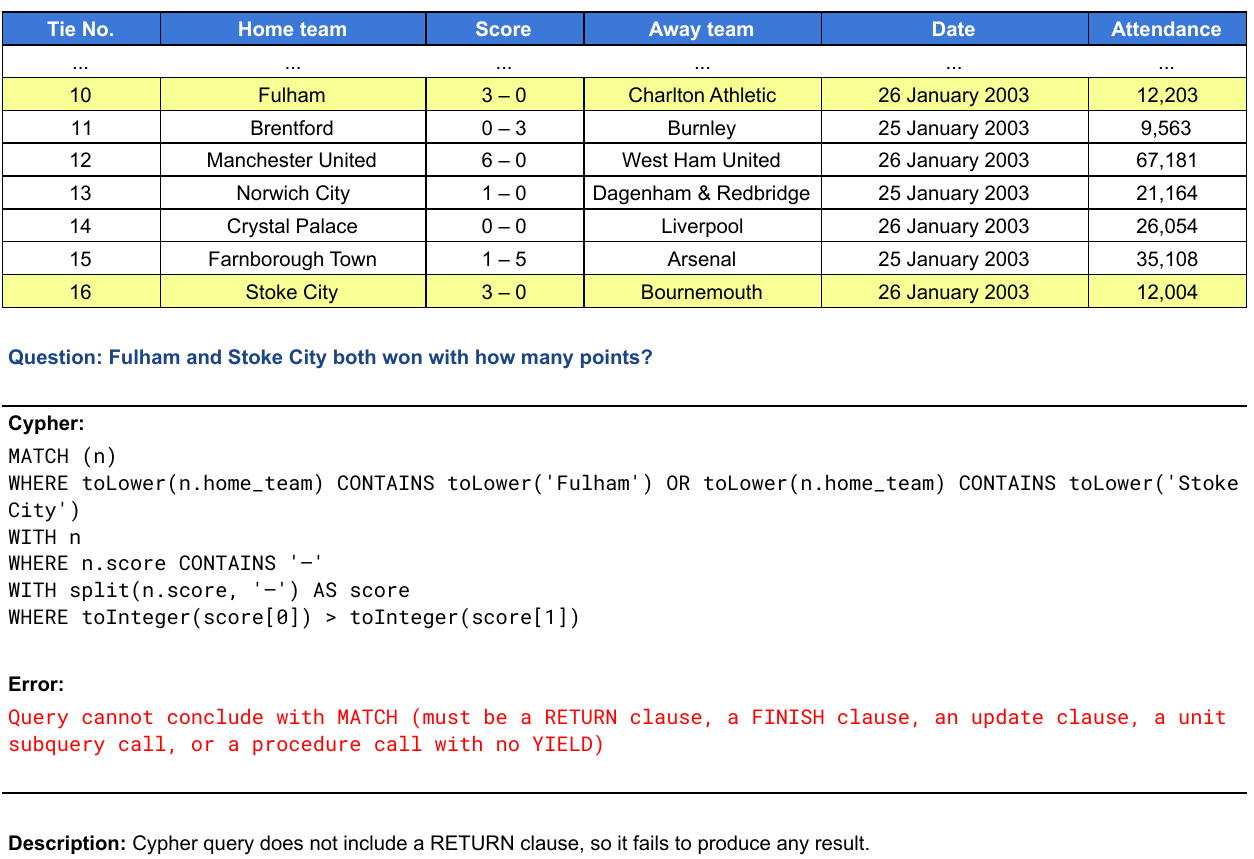}
    \caption{Illustration of Insufficient entity retrieval error.}
    \label{fig:no-answer-cypher-error}
\end{figure*}

Figure~\ref{fig:fullttext-search-error} and Figure~\ref{fig:no-answer-cypher-error} illustrate insufficient entity retrieval errors.
\label{sec:error-search-insufficient}

\subsubsection{Incorrect self-generated entity error}
\label{sec:error-search-incorrect}

\begin{figure*}
    \centering
    \includegraphics[width=\linewidth]{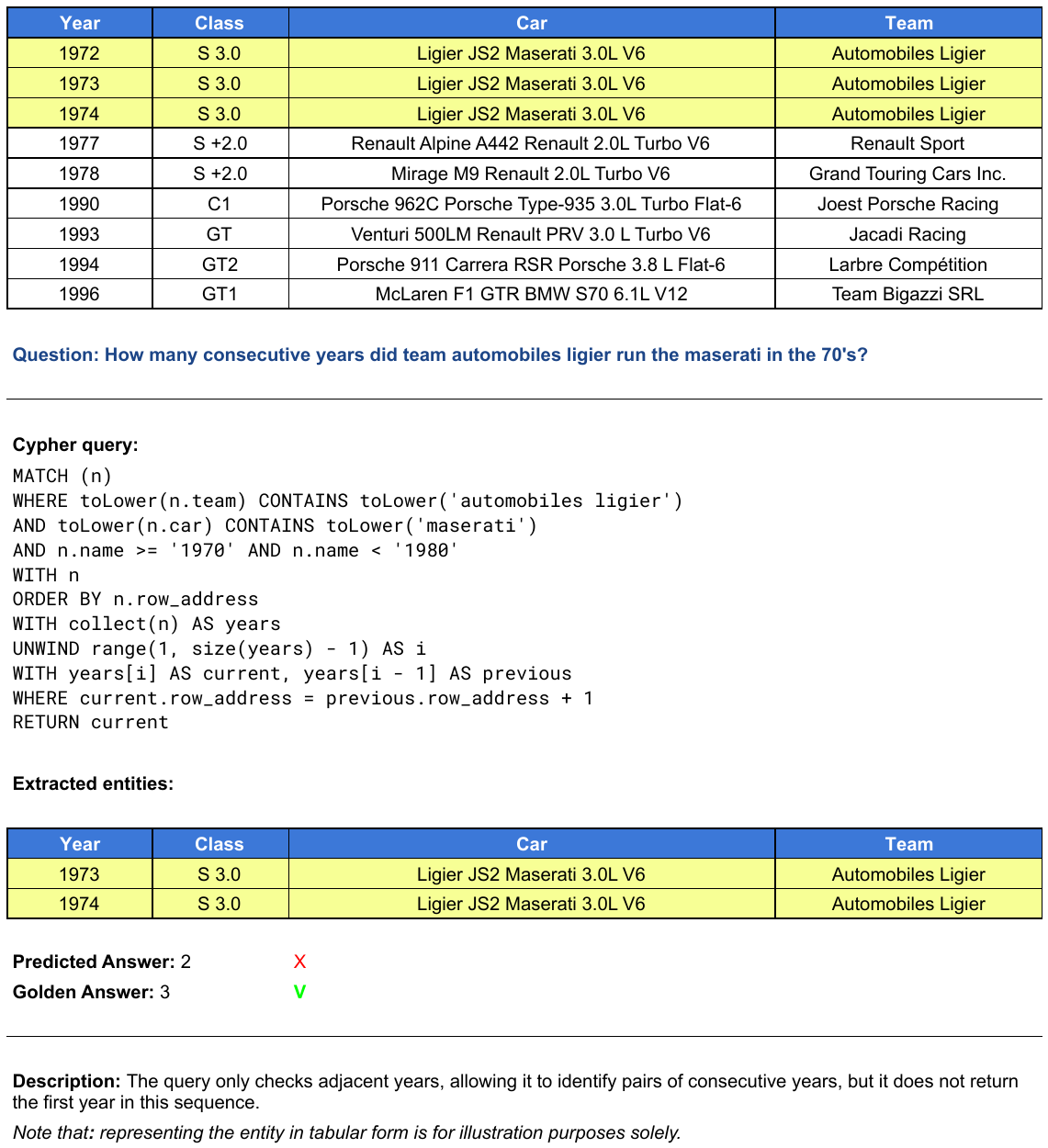}
    \caption{Illustration of Incorrect self-generated entity error.}
    \label{fig:incorrect-cypher-error}
\end{figure*}

Figure~\ref{fig:incorrect-cypher-error} illustrates an incorrect self-generated entity error.

\subsection{Answer generation errors}

\subsubsection{Comparative error}
\label{sec:error-answer-comparative}

\begin{figure*}
    \centering
    \includegraphics[width=\linewidth]{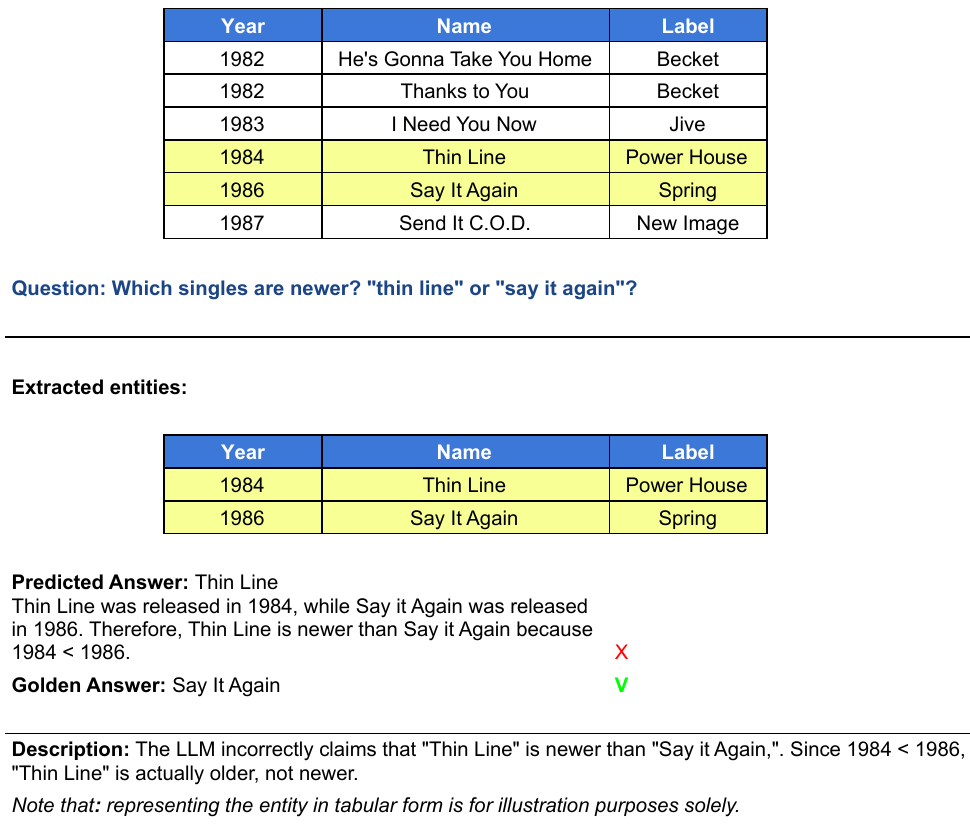}
    \caption{Illustration of Comparative error.}
    \label{fig:comparative-error}
\end{figure*}

Figure~\ref{fig:comparative-error} illustrates a comparative error.

\subsubsection{Numerical error}
\label{sec:error-answer-numeric}

\begin{figure*}
    \centering
    \includegraphics[width=\linewidth]{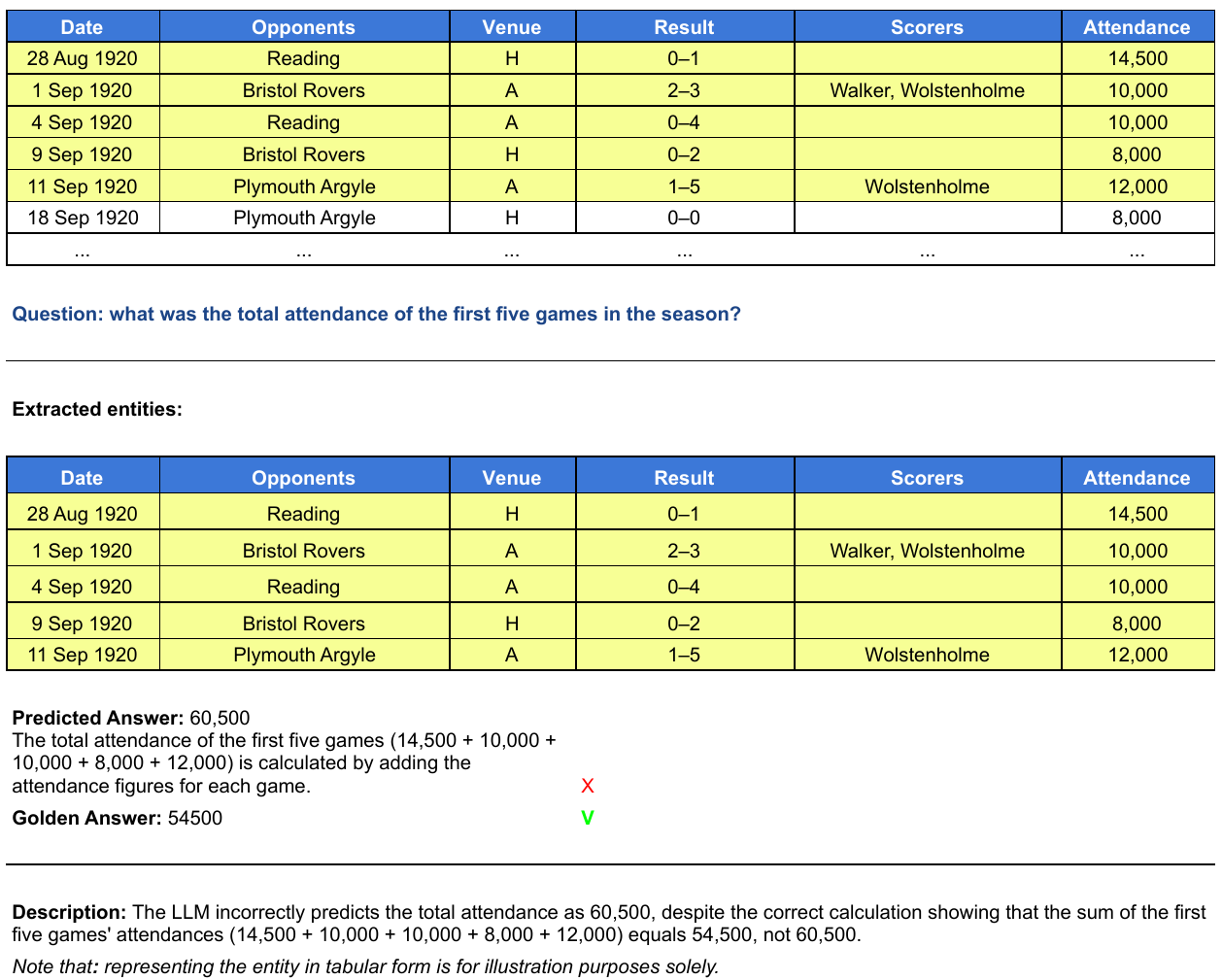}
    \caption{Illustration of Numerical error.}
    \label{fig:numerical-error}
\end{figure*}

Figure~\ref{fig:numerical-error} illustrates a numerical error.

\subsubsection{Logical error}
\label{sec:error-answer-logical}

\begin{figure*}
    \centering
    \includegraphics[width=\linewidth]{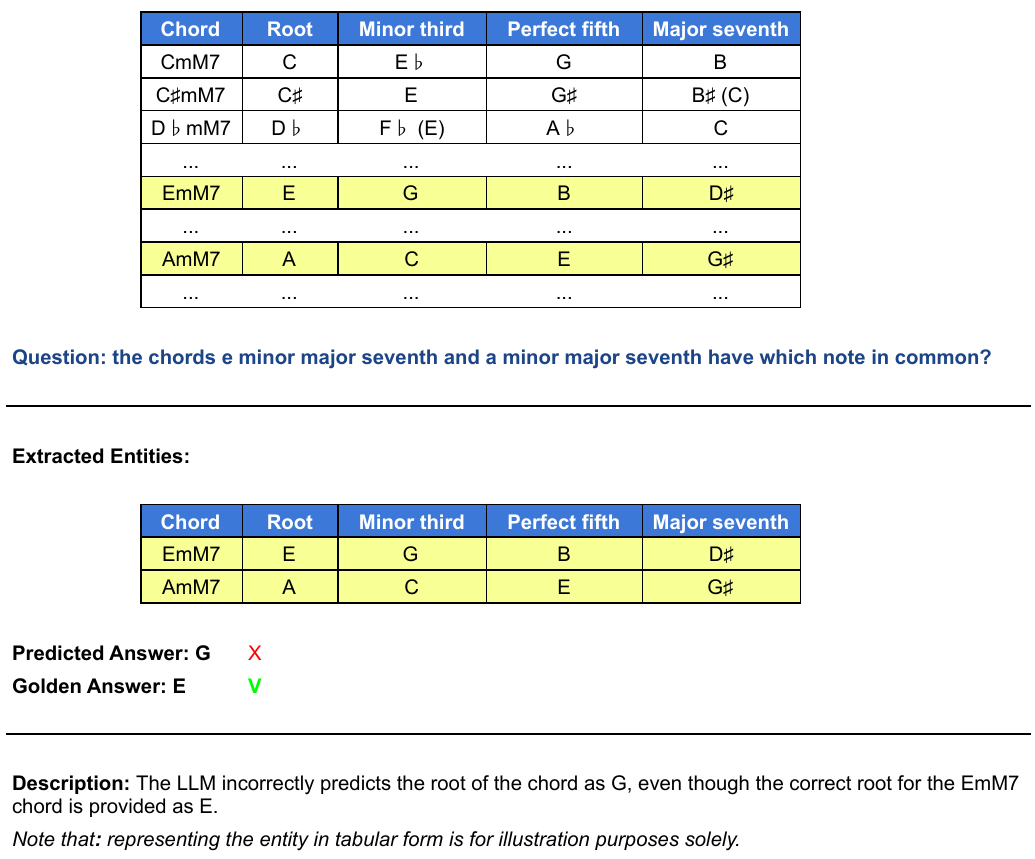}
    \caption{Illustration of Logical error.}
    \label{fig:logical-error}
\end{figure*}

Figure~\ref{fig:logical-error} illustrates a logical error.

\end{document}